\documentclass[conference,a4paper]{IEEEtran}
\IEEEoverridecommandlockouts
\usepackage{cite}
\usepackage{amsmath,amssymb,amsfonts}
\usepackage{dsfont}

\usepackage{algorithm}
\usepackage{algorithmic}

\usepackage{graphicx}
\usepackage{textcomp}
\usepackage{xcolor}

\usepackage{booktabs}
\def\BibTeX{{\rm B\kern-.05em{\sc i\kern-.025em b}\kern-.08em
    T\kern-.1667em\lower.7ex\hbox{E}\kern-.125emX}}

\begin{document}

\title{Adjacency-Based Spectral Proxy Control of Mobile Communication Agents}

\author{
\IEEEauthorblockN{Mariana del Castillo}
\IEEEauthorblockA{\textit{Facultad de Ingeniería} \\
\textit{Universidad de la República}\\
Montevideo, Uruguay \\
mdelcastillo@fing.edu.uy}
\and
\IEEEauthorblockN{Federico Larroca}
\IEEEauthorblockA{\textit{Facultad de Ingeniería} \\
\textit{Universidad de la República}\\
Montevideo, Uruguay \\
flarroca@fing.edu.uy}
}

\maketitle

\begin{abstract}

We consider a heterogeneous mobile-agent network composed of uncontrolled task agents and controllable communication agents. The objective is to reposition communication agents online as task agents move. Since throughput-based objectives are generally unsuitable for real-time control, spectral graph metrics such as algebraic connectivity are commonly adopted as surrogate objectives. 
However, controlling algebraic connectivity relies on the eigenvector corresponding to the second-smallest eigenvalue of a graph's Laplacian matrix (i.e., the Fiedler vector), whose distributed estimation requires an unbounded number of communication rounds to converge.
In this work, we identify a structural decomposition of this Fiedler-gradient controller into a local interaction rule and a graph embedding component, suggesting the use of alternative embeddings that are easier to estimate distributively than the Fiedler vector. As a particular instance, we propose A-Fiedler, which replaces the Fiedler embedding with the dominant eigenvector of the adjacency matrix, commonly used as a graph embedding of nodes into a latent geometry. This representation is more naturally suited for distributed implementation under local communication constraints.
We evaluate A-Fiedler against the classical Fiedler-gradient controller. Results show comparable network performance in the absence of communication constraints and improved robustness under distributed estimation. For instance, under the same number of communication rounds, the Fielder-gradient may even converge to disconnected configurations whereas our proposition maintains performance. We believe our contribution provides a simpler path toward distributed network control.
\end{abstract}


\section{Introduction}
\label{sec:intro}
Maintaining reliable communication in networks of heterogeneous mobile agents is a fundamental challenge in multi-robot applications such as disaster response and environmental monitoring~\cite{qadir2021addressing,schofield2019drones}. A widely used architecture consists of \emph{task agents}, whose motion is dictated by the mission, and \emph{communication agents}, whose positions can be controlled to support communication among task agents~\cite{AnetworkMultiagent, GNN-throughput, egnntopology}. This paper addresses the online repositioning of communication agents with the goal of improving network performance through distributed control.

Directly optimizing network performance is generally unsuitable for online control due to its computational cost and lack of a tractable gradient. A common alternative is to maximize the algebraic connectivity of the communication graph, given by the second-smallest eigenvalue of the graph Laplacian (the Fiedler value)~\cite{fiedler1973algebraic,kim2005maximizing}. Although the resulting gradient-based controller is typically derived assuming centralized knowledge of the graph, its update decomposes into local interactions between neighboring agents, suggesting a distributed implementation.

The main challenge lies in the spectral information required by the controller. In particular, the gradient depends on the Fiedler eigenvector, the eigenvector associated with the Fiedler value, which is a global graph quantity. While distributed estimation methods based on consensus and power iteration exist~\cite{zavlanos2008distributed}, their accuracy depends on the number of communication rounds available between controller updates. In dynamic scenarios, where communication opportunities are limited, the resulting estimation errors may fundamentally alter the control law being implemented, with a substantial impact on network performance, even converging to disconnected configurations, as demonstrated in Section~\ref{sec:results}.

In this work, we identify a structural decomposition underlying the Fiedler-gradient controller. Specifically, we show that the gradient can be separated into a local interaction rule determined by the communication model and a spectral embedding component determined by the graph. This observation enables the design of alternative controllers that preserve the local interaction mechanism while replacing the original embedding with representations better suited for distributed implementation.

As a particular instance of this idea, we investigate the dominant eigenvector of the weighted adjacency matrix $\mathbf{A}$, which, like the Fiedler eigenvector, provides a spectral embedding of graph nodes~\cite{sussman2012consistent,athreya2018statistical}. This choice leads to the \emph{A-Fiedler} controller, whose embedding can be estimated distributively with deterministic convergence guarantees, thereby avoiding the accumulation of approximation errors caused by incomplete consensus.
We evaluate the proposed approach against the Fiedler-gradient controller, showing that it preserves network performance when accurate spectral information is available while providing significantly improved robustness under communication-limited spectral estimation.
\section{System Model and Problem Statement}
\label{sec:model}

We consider $N = N_T + N_C$ agents in $\mathbb{R}^2$. The first $N_T$ agents are \emph{task agents}, with positions $\mathbf{x}_T = \{x_i \in \mathbb{R}^2 : i = 1,\ldots,N_T\}$ that evolve under external dynamics; from the optimization's perspective, they are a time-varying, uncontrolled input. The remaining $N_C$ agents are \emph{communication agents}, with positions $\mathbf{x}_C = \{x_j \in \mathbb{R}^2 : j = N_T+1,\ldots,N_T+N_C\}$ that are the sole decision variables.
 
Between every pair of agents $(i,j)$, the achievable communication rate $C_{ij}$ is a decreasing function of their distance~\cite{fink2011communication}. A link exists between agents $i$ and $j$ whenever $C_{ij}$ exceeds a threshold $C_{\min}$. 
The resulting weighted adjacency matrix $\mathbf{A}$ and graph Laplacian $\mathbf{L}$ are given by
\begin{equation}
  A_{ij} = C_{ij}\,\mathds{1}{\{C_{ij} > C_{\min}\}}, \qquad \mathbf{L} = \mathbf{D} - \mathbf{A},
  \label{eq:adjlap}
\end{equation}
where $\mathbf{D} = \mathrm{diag}(\mathbf{A}\mathbf{1})$ is the diagonal matrix of weighted node degrees.
 
As discussed in Section~\ref{sec:intro}, we adopt the standard approach of maximizing the algebraic connectivity of the communication graph, $\lambda_2(\mathbf{L})$, the second-smallest eigenvalue of $\mathbf{L}$, as a tractable proxy for network performance:
\begin{equation}
  \max_{\mathbf{x}_C} \ \lambda_2\big(\mathbf{L}(\mathbf{x}_T, \mathbf{x}_C)\big).
  \label{eq:problem}
\end{equation}
Because $\mathbf{x}_T$ varies continuously, this proxy problem is itself solved via gradient ascent rather than as a one-shot optimization, re-estimating and applying the resulting update online as the network evolves. The derivation of this gradient is presented in the following section.
 
\subsection{Gradient of the Fiedler Value}
 
Let $\mathbf{v_2}$ be the unit-norm eigenvector associated with the algebraic connectivity $\lambda_2$. We first compute the sensitivity of $\lambda_2$ with respect to the weight $C_{ij}$ of a single edge $(i,j)$, while holding all other edge weights fixed.
Differentiating the eigenvalue equation $\mathbf{L}\mathbf{v_2}=\lambda_2\mathbf{v_2}$ with respect to $C_{ij}$ gives
\begin{equation}
    \frac{\partial\mathbf{L}}{\partial C_{ij}}\mathbf{v_2}
    +  \mathbf{L}\frac{\partial \mathbf{v_2}}{\partial C_{ij}}
    =  \frac{\partial\lambda_2}{\partial C_{ij}}\mathbf{v_2}
    +  \lambda_2\frac{\partial\mathbf{ v_2}}{\partial C_{ij}}.
\end{equation}
Left-multiplying by $\mathbf{v}_2^T$ and using the symmetry of $\mathbf{L}$ and the normalization condition on $\mathbf{v}_2$, the terms involving $\partial \mathbf{v_2}/\partial C{ij}$ cancel, yielding
\begin{equation}
    \frac{\partial\lambda_2}{\partial C_{ij}}
    =
    \mathbf{v_2}^T
    \frac{\partial\mathbf{L}}{\partial C_{ij}}
    \mathbf{v_2}.
\end{equation}
Perturbing $C_{ij}$ affects four entries of $\mathbf{L}$ (the diagonals in $i$ and $j$ and the corresponding entries $ij$ and $ji$), which can be made explicit through the incidence decomposition of the Laplacian matrix to obtain
\begin{equation}
\begin{aligned}
\frac{\partial\lambda_2}{\partial C_{ij}} &=
 \mathbf{v_2}_i^2+ \mathbf{v_2}_j^2-2 \mathbf{v_2}_i \mathbf{v_2}_j\\
&= ( \mathbf{v_2}_i- \mathbf{v_2}_j)^2,
\end{aligned}
\end{equation}
where $\mathbf{v_2}_i$ corresponds to the $i$-th entry of the Fiedler vector. Applying the chain rule to this last equality yields the gradient with respect to $x_i$,
\begin{equation}
    \nabla_{x_i}\lambda_2  =  \sum_{j\in\mathcal N(i)}  \nabla_{x_i}C(\|x_i-x_j\|)\,(\mathbf{v_2}_i-\mathbf{v_2}_j)^2,
    \label{eq:fiedler_grad}
\end{equation}
where $\mathcal N(i)$ denotes the set of neighbors of agent $i$, and $\nabla_{x_i}C(\|x_i-x_j\|)$ is the gradient of the channel model with respect to $x_i$:
\begin{equation}
    \nabla_{x_i} C(\|x_i-x_j\|)=C'(\|x_i-x_j\|)\frac{(x_i-x_j)}{\|x_i-x_j\|}.
\end{equation}
Equation~\eqref{eq:fiedler_grad} shows that the control action is obtained as a weighted sum of local interaction vectors $\nabla_{x_i}C_{ij}$. The contribution of each neighbor is scaled by the squared distance $(\mathbf {v_2}_i-\mathbf {v_2}_j)^2$ between the corresponding nodes in the one-dimensional latent geometry induced by the Fiedler embedding. Consequently, the controller combines local motion directions with weights determined by distances in a global latent representation of the communication graph. While the interaction rule itself is local, obtaining these spectral weights requires estimating a global graph quantity, which is the main challenge for distributed implementation disccused next.

\section{Distributed Implementation}
\label{sec:distributed}

The Fiedler-gradient controller can be viewed as a particular instance of the following embedding-based interaction rule:
\begin{equation}
x_i(k+1)=x_i(k)+\alpha_m g_i(k),
\label{eq:outer_update}
\end{equation}
where $g_i(k)$ is the gradient direction computed from the current graph embedding:
\begin{equation}
g_i(k)=
\sum_{j\in\mathcal N(i)}
\nabla_{x_i}C(\|x_i-x_j\|)
\,d_{\phi}^2(i,j).
\end{equation}

In this formulation, $d_{\phi}(i,j)$ denotes the distance between agents $i$ and $j$ in the latent geometry induced by the embedding $\phi$.
For the classical Fiedler-gradient controller, this embedding is given by $\phi(i)=\mathbf {v_2}_i$, where $\mathbf {v_2}$ is the Fiedler eigenvector. Estimating this embedding in a distributed manner is the main implementation challenge. The standard approach for distributed eigenvector estimation is power iteration, which relies on local matrix-vector multiplications over the graph. However, power iteration converges to the eigenvector associated with the largest eigenvalue, whereas the Fiedler vector corresponds to the second-smallest eigenvalue of the Laplacian matrix. Therefore, two modifications are required before applying this method. First, the Laplacian operator is shifted and scaled so that the Fiedler eigenvector becomes a dominant component of the iteration. Second, the eigenvector associated with the zero eigenvalue must be removed through deflation. This is implemented by subtracting the average value of the estimated vector at each iteration, as described in Algorithm~\ref{alg:l_dist}~\cite{yang2010decentralized}. 

\begin{algorithm}[b]
\caption{L-Fiedler-distributed: distributed estimation of the Fiedler coordinate $\mathbf{v_2}_{i}$}\label{alg:l_dist}
\begin{algorithmic}[1]
\STATE $d_{\max} \leftarrow \textsc{MaxConsensus}(D_{ii};\,T_{\max}\text{ rounds})$
\STATE $\varepsilon \leftarrow 0.8/(2d_{\max})$ 
\COMMENT{\emph{using the bound $\lambda_{\max}(\mathbf L)\leq 2d_{\max}$}}
\FOR{$t=1$ to $T_{\mathrm{pow}}$}
    \STATE $\mathbf {v_2}_i \leftarrow \mathbf {v_2}_i-\varepsilon\Big(D_{ii}\mathbf {v_2}_i-\sum_{j\in\mathcal N(i)}C_{ij}\mathbf {v_2}_j\Big)$
    \STATE $\bar{\mathbf{v_2}} \leftarrow \textsc{AvgConsensus}(\mathbf{v_2}_i;\,T_{\mathrm{avg}}\text{ rounds})$
    \STATE $\mathbf {v_2}_i \leftarrow \mathbf {v_2}_i-\bar{\mathbf {v_2}}$
    \STATE $r \leftarrow \textsc{AvgConsensus}(\mathbf {v_2}_i^2;\,T_{\mathrm{avg}}\text{ rounds})$
    \STATE $\mathbf {v_2}_i \leftarrow \mathbf {v_2}_i/\sqrt{r}$
\ENDFOR
\RETURN $\mathbf {v_2}_i$
\end{algorithmic}
\end{algorithm}

The proposed distributed implementation relies on consensus algorithms only for the auxiliary global quantities required by the spectral estimation procedure. In particular, three consensus operations are used: a maximum consensus computes the maximum weighted degree $d_{\max}$ required to select the scaling factor $\varepsilon$, while average consensus is used twice per iteration, to remove the component of the estimated vector along the trivial eigenvector $\mathbf{1}$ and to normalize its norm. Both primitives require only neighbor-to-neighbor exchanges over the communication graph and perform, for $T$ rounds, the updates
\begin{align}
  \textsc{MaxConsensus}: \quad
    & z_i \leftarrow \max\Big(z_i,\; \max_{j\in\mathcal N(i)} z_j\Big),
    \label{eq:maxcons}\\
  \textsc{AvgConsensus}: \quad
    & z_i \leftarrow w_{ii}\, z_i + \sum_{j\in\mathcal N(i)} w_{ij}\, z_j ,
    \label{eq:avgcons}
\end{align}
where the weights in \eqref{eq:avgcons} are symmetric, $w_{ij}=w_{ji}$, and satisfy $w_{ii} + \sum_{j\in\mathcal N(i)} w_{ij}=1$, so that the iteration preserves the network sum and converges geometrically to the average $\tfrac{1}{n}\sum_i z_i$ \cite{xiao2004fast}.
The two primitives differ fundamentally in their convergence behavior. The iteration \eqref{eq:maxcons} terminates exactly, yielding the global maximum within a number of rounds bounded by the graph diameter, whereas \eqref{eq:avgcons} converges only asymptotically~\cite{olfatisaber2007consensus}: after $T$ rounds the residual error is of order $\rho^{T}$, with $\rho<1$ the second largest eigenvalue modulus of the weight matrix, guaranteed since the graph is connected and $w_{ii}>0$ for all $i$~\cite{xiao2004fast}. Under a limited communication budget, these residuals perturb the estimated Fiedler embedding and propagate to successive controller updates, potentially leading to inaccurate control actions and, in challenging scenarios, network disconnection.
This limitation motivates the A-Fiedler controller, introduced next.

\subsection{Alternative Distributed Embedding}

The previous analysis shows that the main difficulty of the Fiedler-gradient controller lies in the distributed estimation of its embedding, rather than in the local interaction rule itself. We thus propose A-Fiedler, which preserves the same interaction structure while replacing the latent geometry induced by the Fiedler embedding. Specifically, the Fiedler latent distance
\begin{equation}
    d_{\mathrm{F}}^2(i,j)=(\mathbf {v_2}_i-\mathbf {v_2}_j)^2
\end{equation}
is replaced by an adjacency-induced latent distance
\begin{equation}
    d_{\mathrm{A}}^2(i,j)=(\mathbf {u}_i-\mathbf {u}_j)^2,
\end{equation}
where $\mathbf{u}$ is the dominant eigenvector of the weighted adjacency matrix $\mathbf A$.

Between consecutive position updates, agents estimate their embedding coordinates through an inner iterative procedure. For the legacy Fiedler-gradient, that would be Algorithm~\ref{alg:l_dist}, whereas Algorithm~\ref{alg:a_dist} is used instead for our proposed A-Fiedler. Since the network topology changes only through the outer-loop motion (i.e., Equation \eqref{eq:outer_update}), each inner loop is initialized with the embedding estimate obtained in the previous outer iteration. The two implementations differ only in the procedure used to estimate these coordinates.

\begin{algorithm}
\caption{A-Fiedler-distributed: distributed estimation of $\mathbf {u}_i$}
\label{alg:a_dist}
\begin{algorithmic}[1]
\REQUIRE neighbor weights $\{C_{ij}\}_{j\in\mathcal N(i)}$, previous estimate $\mathbf {u}_i$, $T_{\mathrm{pow}}$, $T_{\max}$
\FOR{$t=1$ to $T_{\mathrm{pow}}$}
    \STATE $\mathbf {u}_i \leftarrow \sum_{j\in\mathcal N(i)} C_{ij}\,\mathbf {u}_j$
    \STATE $m \leftarrow \textsc{MaxConsensus}(|\mathbf {u}_i|;\,T_{\max}\text{ rounds})$
    \STATE $\mathbf {u}_i \leftarrow \mathbf {u}_i/m$
\ENDFOR
\RETURN $\mathbf {u}_i$
\end{algorithmic}
\end{algorithm}

Note that the implementations are parameterized by three budgets: $T_{\mathrm{pow}}$ power-iteration steps, $T_{\mathrm{avg}}$ communication rounds for each average-consensus call, and $T_{\max}$ rounds for each max-consensus call. Each power iteration consists of one local matrix--vector multiplication implemented by message passing followed by a scaling step, whereas the Laplacian implementation additionally removes the component along the trivial eigenvector. 

The key difference between the implementations lies in the consensus operations required to estimate the embedding. A-Fiedler requires only max consensus for eigenvector scaling, which converges exactly in finite time. In contrast, L-Fiedler requires average consensus both for deflation and normalization. Since average consensus converges only asymptotically, finite communication budgets result in approximate Fiedler coordinates. This difference explains the larger performance gap between the centralized and distributed Laplacian controllers we will observe in the next section.

\section{Experiments and Results}
\label{sec:results}

\subsection{Static experiments}

We evaluate the proposed approach by separating the effect of the proposed spectral embedding from the impact of distributed spectral estimation. To this end, we compare implementations using exact spectral coordinates with implementations based on local spectral estimation.

The evaluation is performed over multiple network realizations. Task agents are placed at random within a region whose area scales with the total number of agents $N$, maintaining approximately constant agent density across network sizes. Communication agents are initialized sequentially at the centroids of the largest triangles of the Delaunay triangulation formed by the agents placed so far. Realizations in which the task-agent graph is already connected are discarded, ensuring that communication agents play an essential role in initial network connectivity.
\begin{table*}[t]
\centering
\caption{Performance comparison for different network sizes. Relative MNF change and convergence statistics are averaged over 20 realizations.}
\label{tab:results}
\begin{tabular}{lccc ccc ccc}
\toprule
& \multicolumn{3}{c}{$N=5$}
& \multicolumn{3}{c}{$N=8$}
& \multicolumn{3}{c}{$N=10$} \\
\cmidrule(lr){2-4}
\cmidrule(lr){5-7}
\cmidrule(lr){8-10}
Method &
Disc. &
MNF &
Conv. &
Disc. &
MNF &
Conv. &
Disc. &
MNF &
Conv. \\
&
(\%) &
(\%) $\pm$ SD &
iter. &
(\%) &
(\%) $\pm$ SD &
iter. &
(\%) &
(\%) $\pm$ SD &
iter. \\
\midrule
A\_exact & 0.0  & $+10.5 \pm 9.1$ & 62.7 & 0.0  & $+5.4 \pm 5.6$ & 73.8 & 0.0  & $+0.7 \pm 3.2$ & 80.8 \\
L\_exact & 0.0  & $+16.1 \pm 6.5$ & 53.0 & 0.0  & $+7.0 \pm 5.1$ & 70.6 & 0.0  & $+1.7 \pm 2.7$ & 80.3 \\
A\_dist  & 0.0  & $+10.5 \pm 9.1$ & 62.7 & 0.0  & $+5.4 \pm 5.6$ & 74.0 & 0.0  & $+0.7 \pm 3.2$ & 80.8 \\
L\_dist  & 95.0 & $-320.6 \pm 131.0$ & 81.5 & 15.0 & $-31.7 \pm 86.1$ & 60.5 & 20.0 & $-25.9 \pm 62.4$ & 76.7 \\
\bottomrule
\end{tabular}
\end{table*}

We evaluate four controller variants based on the two spectral embeddings, denoted $L_{\mathrm{exact}}$, $A_{\mathrm{exact}}$, $L_{\mathrm{dist}}$, and $A_{\mathrm{dist}}$. The exact implementations use the corresponding eigenvectors directly (simulating an unbounded communication budget), whereas the distributed implementations estimate these embeddings using the procedures described in Section~\ref{sec:distributed}. Since the two embeddings produce gradients with different scales, the step size $\alpha_m$ is calibrated separately for each method so that all controllers produce the same initial displacement.

For the distributed implementations, the estimation procedures use the parameters $T_{\mathrm{pow}}$, $T_{\mathrm{avg}}$, and $T_{\max}$ described in Section~\ref{sec:distributed}. The average-consensus horizon for L-Fiedler is set to $T_{\mathrm{avg}}=\lceil (N-1)/2 \rceil$, which provides a communication budget comparable to that of A-Fiedler. For each network size, 20 independent network realizations are generated, and all four controllers are evaluated from the same initial conditions.

The performance metric is the relative change in the multi-commodity network flow (MNF) metric~\cite{egnntopology} with respect to the initial network configuration. This metric captures the overall network performance while accounting for shared-access constraints. Runs that become disconnected are included in the statistics; for these cases, the MNF value used for evaluation corresponds to the last value obtained before disconnection.

Table~\ref{tab:results} summarizes the results obtained across the four controller variants. The exact implementations show that replacing the Fiedler embedding with the adjacency-based embedding leads to only a modest reduction in MNF improvement across all tested network sizes. Moreover, the distributed implementation of A-Fiedler closely matches its exact counterpart, indicating that the proposed embedding is robust to distributed estimation.

In contrast, the distributed L-Fiedler controller exhibits a substantial degradation with respect to $L_{\mathrm{exact}}$, including frequent disconnections. This results in negative average MNF changes despite the strong performance of the exact Laplacian-based controller. These results indicate that the main limitation of the distributed L-Fiedler implementation is not the local interaction rule, but the accuracy with which its spectral information can be estimated under limited communication constraints.
Under comparable communication budgets, the performance difference is primarily attributable to the spectral estimation procedure. A-Fiedler relies on finite-time max-consensus normalization, whereas L-Fiedler requires deflation and normalization through truncated average consensus.

\subsection{Dynamic experiments}

\begin{figure*}
\centering
\includegraphics[width=0.96
\textwidth]{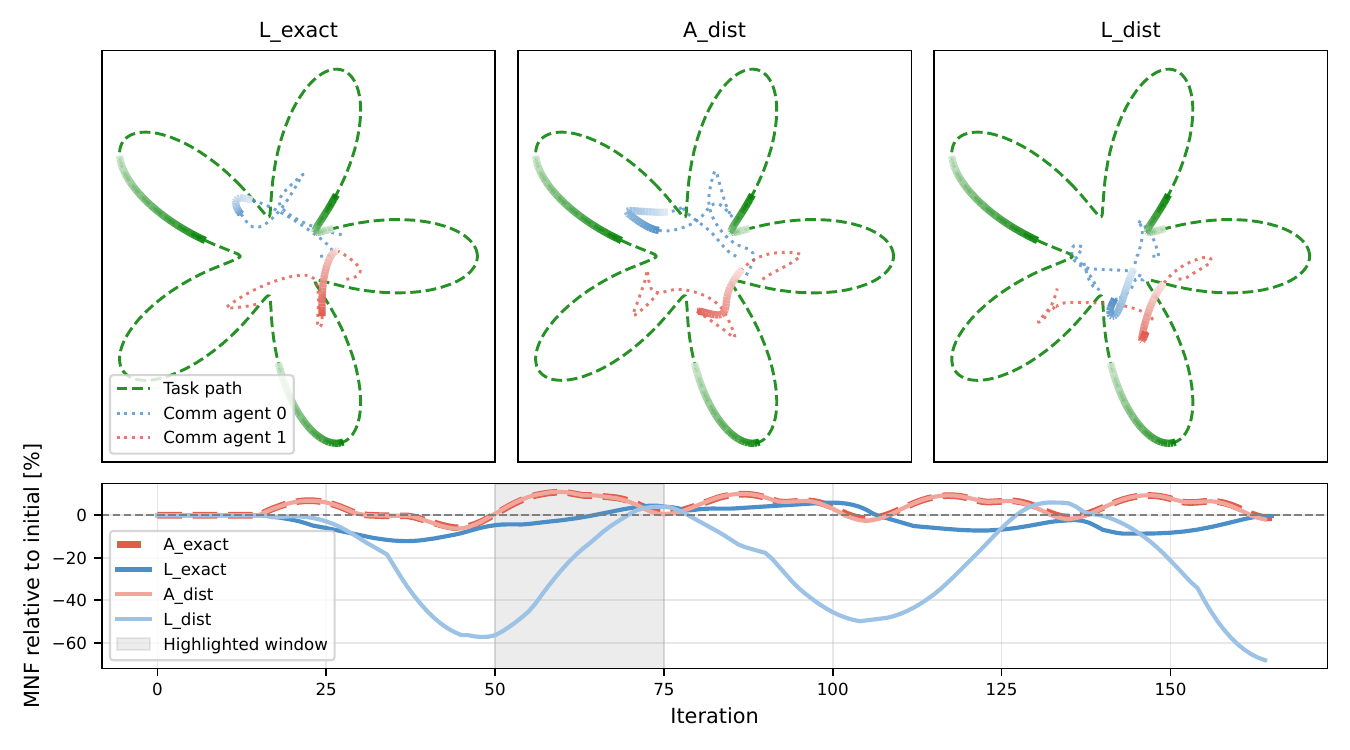}
\caption{Dynamic tracking experiment comparing the four methods. Top: communication-agent repositioning while task agents follow a clover-shaped trajectory. Bottom: relative MNF evolution over time. The highlighted window indicates the time interval corresponding to the bold trajectory segments shown in the top panels.}
\label{fig:trebol}
\end{figure*}

The previous experiments evaluate the controllers over multiple independent network realizations where the task agents were static. We next illustrate their online behavior in a dynamic
scenario where task agents follow a prescribed clover-shaped trajectory; see Fig.\ \ref{fig:trebol}. The trajectory is unknown to the controllers and is selected to repeatedly modify the relative geometry between task agents, creating varying communication conditions. 

For visual clarity, the trajectory of $A_{\mathrm{exact}}$ is omitted from the
figure since it overlaps almost completely with that of $A_{\mathrm{dist}}$. The bottom panel confirms that both implementations exhibit nearly identical MNF evolution, even if the latter uses a fully distributed estimation of the embeddings, maintaining the MNF metric within approximately $\pm10\%$ of its initial value throughout the experiment. In contrast, $L_{\mathrm{dist}}$ presents large oscillations and reaches a degradation of nearly 70\% in relative MNF, while $L_{\mathrm{exact}}$ exhibits intermediate behavior. This example illustrates the ability of the proposed controller to maintain stable operation as the communication topology evolves.

\section{Conclusion}
\label{sec:conclusion}

This work identified a structural decomposition of the classical Fiedler-gradient controller into a local interaction rule and a latent geometry that defines pairwise interaction weights. This observation separates the communication model from the latent representation used to weight local interactions, revealing that the Fiedler embedding is only one possible choice.

Based on this observation, we proposed A-Fiedler, which replaces the Fiedler latent geometry with one induced by the dominant eigenvector of the weighted adjacency matrix. Unlike the classical Fiedler-gradient controller, A-Fiedler does not directly optimize the algebraic connectivity objective $\lambda_2(\mathbf{L})$. Instead, it provides a spectral proxy that is simpler to estimate under communication constraints. The experimental results show that this approximation introduces only a modest performance loss when exact spectral information is available, while significantly improving robustness in distributed estimation.

The difference between the two distributed implementations is mainly explained by the accuracy of the estimated spectral coordinates under finite communication budgets. A-Fiedler relies on finite-time max-consensus normalization, whereas L-Fiedler requires approximate average consensus for deflation and normalization. The influence of the estimation horizon, consensus accuracy, and update parameters on controller stability and scalability remains an open question.

More broadly, this formulation suggests a family of latent-geometry controllers rather than a single Fiedler-based design. The adjacency embedding investigated here represents only one possible realization, and other latent representations, including community-oriented spectral embeddings or learned graph representations, could be explored while preserving the same local interaction mechanism.
\section*{Acknowledgment}
The authors would like to thank Ing. Santiago Fernandez for his fresh perspective that helped shape the direction of this work.

\bibliographystyle{IEEEtran}
\bibliography{biblo}

\end{document}